\documentclass[runningheads]{llncs}

\usepackage{eccv}

\usepackage{eccvabbrv}

\usepackage{graphicx}
\usepackage{booktabs}

\usepackage[accsupp]{axessibility}  

\usepackage{hyperref}

\usepackage{orcidlink}

\begin{document}

\title{Affective Behavior Analysis using Task-adaptive and AU-assisted Graph Network} 

\titlerunning{Abbreviated paper title}

\author{Xiaodong Li \and Wenchao Du \and Hongyu Yang}

\authorrunning{Li et al.}

\institute{School of Computer Science, Sichuan University, China \\
\email{lixiaodong@stu.scu.edu.cn} \\
\email{\{wengchaodu.cs, yanghongyu\}@scu.edu.cn}}

\maketitle

\begin{abstract}
  In this paper, we present our solution and experiment result for the Multi-Task Learning Challenge of the 7th Affective Behavior Analysis in-the-wild(ABAW7) Competition. This challenge consists of three tasks: action unit detection, facial expression recognition, and valance-arousal estimation. We address the research problems of this challenge from three aspects: 1)For learning robust visual feature representations, we introduce the pre-trained large model Dinov2. 2) To adaptively extract the required features of eack task, we design a task-adaptive block that performs cross-attention between a set of learnable query vectors and pre-extracted features. 3) By proposing the AU-assisted Graph Convolutional Network(AU-GCN), we make full use of the correlation information between AUs to assist in solving the EXPR and VA tasks. Finally, we achieve the evaluation measure of \textbf{1.2542} on the validation set provided by the organizers.
\end{abstract}

\section{Introduction}
\label{sec:intro}
Affective computing aims to develop systems with the ability to recognize, interpret, process, and simulate human emotions. And more and more people pay attention to affective computing due to its wide application scenarios. With the development of computer vision and deep learning, there are also numerous studies and modern applications based on deep learning in the field of affective computing \cite{kollias2019deep, kollias2019expression, kollias2019face, kollias2020analysing, kollias2021affect, kollias2021analysing, kollias2021distribution, kollias2022abaw, kollias2023abaw, kollias2023abaw2, kollias2023multi, kollias20246th, zafeiriou2017aff}. In the ABAW7 competition \cite{kollias20247th}, we took part in the multi-task learning(MTL) challenge. The MTL challenge consists of three subtasks: action unit (AU) detection, facial expression recognition(EXPR) and valence-arousal(VA) estimation. 

Our proposed method for MTL Chanllenge consists of four components.

\textbf{1)} We employ pre-trained large model (i.e. Dinov2\cite{oquab2023dinov2}) to learn robust visual features for downstream tasks.

\textbf{2)} We design a task-adaptive block to capture feature of each task from pre-extracted features. 

\textbf{3)} We also propose a AU-assisted Graph Convolutional Network(AU-GCN) to improve performance of EXPR and VA tasks.

\textbf{4)} We achieve a performance index of \textbf{1.2542} on the official validation set.

\section{Methods}
\begin{figure}[t]
   \centering
   \includegraphics[width=\linewidth]{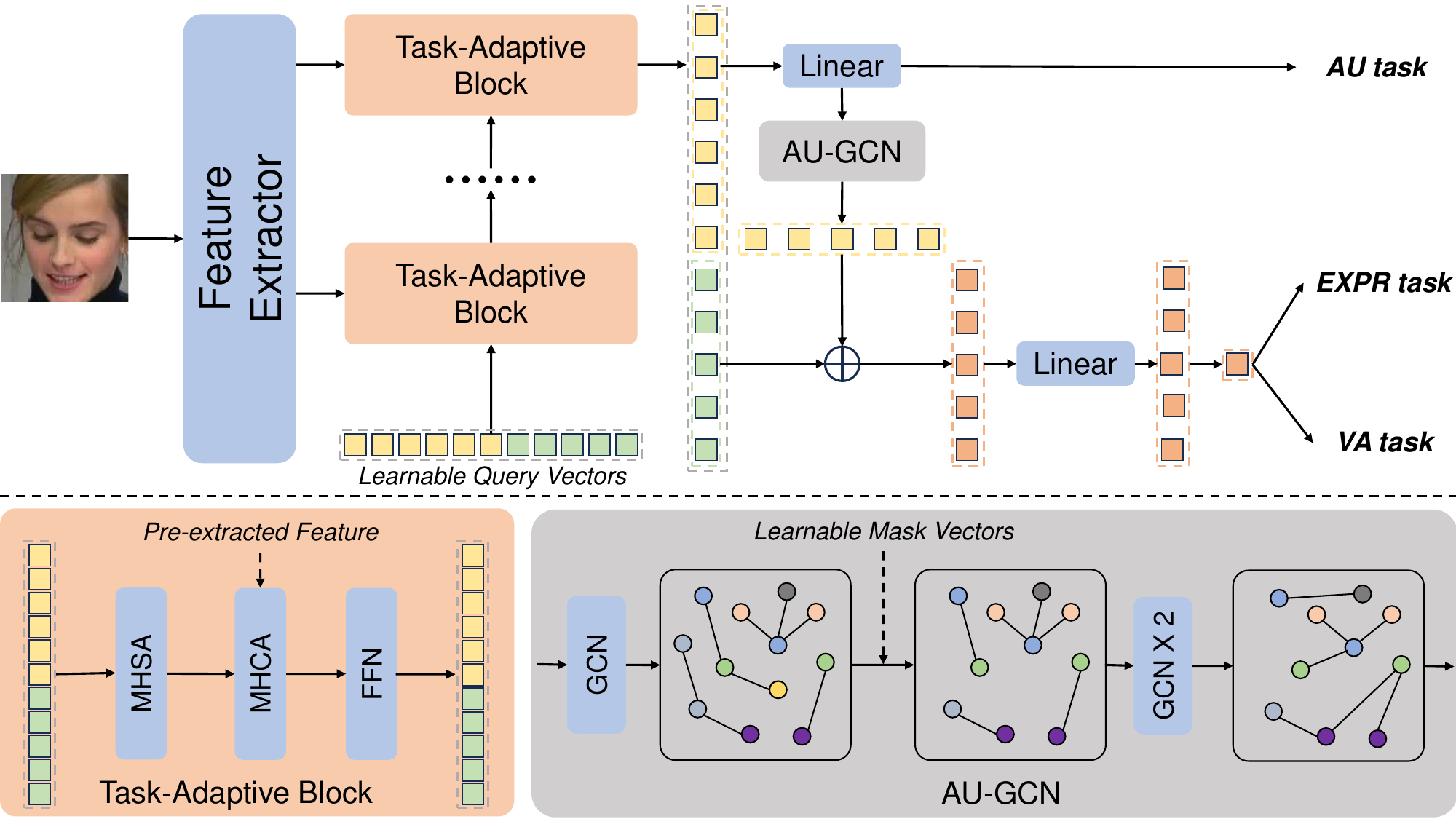}
   \caption{The pipline of our method for the MTL challenge.}
   \label{figure1}
\end{figure}

Given an face image sequence $\mathcal{X} = \{ x_i \in \mathbb{R}^{\textit{C} \times \textit{H} \times \textit{W}}, i = 1, 2, ..., N \}$ and corresponding labels $\mathcal{Y} = \{ y_{va}^i\in \mathbb{R}^2 , y_{expr}^i \in \mathbb{R}^8 , y_{au}^i \in \mathbb{R}^{12} , i = 1, 2, ..., N \}$, where $y_{va}^i$, $y_{expr}^i$ and $y_{au}^i$ denote VA labels, EXPR labels and AU labels, respectively. $y_{va}^i$ is a two-dimensional vector of valence and arousal with the range between [-1, 1]. $y_{expr}^i$ is an integer from 0 to 7, indicating one of the eight emotional categories, i.e., neutral, anger, disgust, fear, happiness, sadness, surprise, and other. $y_{au}^i$ is either 0 or 1, including 12 AU labels, i.e., AU1, AU2, AU4, AU6, AU7, AU10, AU12, AU15, AU23, AU24, AU25 and AU26. Please note that when the values of $y_{va}^i$, $y_{expr}^i$ and $y_{au}^i$ is -5, -1 and -1, the label is considered invalid. Our pipline for MTL Challenge is shown in Fig. \ref{figure1}. 

\subsection{Feature Extraction}
We first resize each image to a size of $238\times238$ and then use the pre-trained large model (i.e. Dinov2) to extract the feature vector $\mathcal{F} \in \mathbb{R}^{289 \times 1536 }$ from the input face image, where ``289'' is the number of patch and ``1536'' means the channel dimension. Dinov2 is trained via self-supervised learning on the LVD-142M datasets, which contains 120 million images. This allows it to generate some universal visual features. Therefore, we decide not to perform further fine-tuning on the official training set. Next, we employ several 1D convolutional layers to compress the channel dimension, resulting in $\mathcal{F'} \in \mathbb{R}^{289 \times 128}$. 

\subsection{Task-Adaptive Block}
To obtain task-adaptive features from $\mathcal{F'}$, we adopt the Query-driven Decoder Moudle proposed by Sun et.al\cite{sun2024taskadaptiveqface}. For three subtasks, we mark it with a set of learnable query vectors $\mathcal{Q} \in \mathbb{R}^{22 \times 128}$, where ``22'' refers the number of queries for all tasks. Specifically, the number of queries for AU, EXPR and VA are 12, 8 and 2, respectively. Then, we stack four task-adaptive blocks, thereby allowing these query vectors to fully select the desired features from $\mathcal{F'}$. As shown in the Fig. \ref{figure1}, each task-adaptive block consists of a multi-head self-attention layer, a multi-head cross-attention layer and a feed-forward network. Considering that the initial query vector is a zero vector, we remove the multi-head self-attention layer in the first task-adaptive block. The task-adaptive block $\textit{i}$ updates the queries $\mathcal{Q}_{i-1}$ as follows:
\begin{align}
\mathcal{Q}_i^1 &= MHSA(\hat{\mathcal{Q}}_{i-1}, \hat{\mathcal{Q}}_{i-1}, \mathcal{Q}_{i-1}), \\
\mathcal{Q}_i^2 &= MHCA(\hat{\mathcal{Q}}_{i}^1, \hat{\mathcal{F'}}, \mathcal{F'}), \\ 
\mathcal{Q}_i &= FFN(\mathcal{Q}_i^2),
\end{align}
where $\hat{*}$ is the vectors modified by adding the position embedding. $\mathcal{Q}_i^1$ and $\mathcal{Q}_i^2$ are two intermediate variables. The MHSA(\textit{q, k, v}), MHCA(\textit{q, k, v}) and FFN($\cdot$) functions are the same as defined in the standard Transformer decoder \cite{vaswani2017attention}. Furthermore, it is worth noting that since the object of the multi-head self-attention layer in task-adaptive block is $\mathcal{Q}$, this block also can model the potential correlations between different tasks.

\subsection{ AU-assisted Graph Convolutional Network}
Inspired by \cite{xie2020assisted} , as shown in Fig. \ref{figure1}, we design AU-assisted Graph Convolutional Network (AU-GCN) to fully leverage the relationships between AUs to assist in solving the EXPR and VA tasks. After obtaining the AU-specific feature vectors $ f_{au} \in \mathbb{R}^{12 \times 128}$ generated by the task-adaptive block, we use a GCN layer to construct a fully connected graph, where each AU represents a node, totaling 12 nodes. Next, we design a set of learnable mask vectors $\mathcal{M} \in \mathbb{R}^{10 \times 128}$ to mask two nodes, making it easier to fuse with the EXPR-specific and VA-specific feature vectors $ f_{expr+va} \in \mathbb{R}^{10 \times 128}$ . Finally, we use two GCNs layer to further enhance the fused features. Following previous work \cite{luo2022learning}, we construct the adjacency matrix used in the GCNs. 

\subsection{Loss function}
For AU task, we utilize the binary cross entropy loss which is given by:
\begin{align}
\mathcal{L}_{au}(\hat{y}_{au}, y_{au}) = -\frac{1}{U} \sum_{i}^{U} P^i_{au} y^i_{au} log(\sigma(\hat{y}_{au})) + (1-y^i_{au}) log(1-\sigma(\hat{y}_{au})),
\end{align}
where U is equal to 12 which denotes the total number of facial action units, $\hat{y}_{au}$ is the AU prediction, $y_{au}$ is the ground truth label for AU, $P^i_{au}$ equals to the number of negative samples divided by the number of positive samples for each AU.

For EXPR task, we utilize the cross entropy loss which is shown in:
\begin{align}
\mathcal{L}_{expr}(\hat{y}_{expr}, y_{expr}) = - \sum_{i}^{C} P^i_{expr} y^i_{expr} log((\hat{y}_{expr})),
\end{align}
where C is 8 which means the number of expression categories, $P^i_{expr}$ is the weight of each expression for data balancing.

For VA task, we utilize the negative Concordance Correlation Coefficient (CCC) as the inference loss:
\begin{align}
\mathcal{L}_{va}(\hat{y}_{va}, y_{va}) = 1 - CCC_V + 1 - CCC_A,
\end{align}

For the MTL task, the optimization objective $\mathcal{L}$ can be expressed as :
\begin{align}
\mathcal{L} = \mathcal{L}_{au} + \mathcal{L}_{expr} + \mathcal{L}_{va},
\end{align}

\section{Experiment}
\subsection{Dataset and Setting}
\textbf{Dataset}. The dataset we used in the MTL challenge is s-Aff-Wild2, which is a static version of Aff-Wild2 database. s-Aff-Wild2 contains around 221k images selected from Aff-Wild2, with annotations in three dimensions: 12 action units, 8 expressions, and 2 continuous emotion labels in valence and arousal. In the s-Aff-Wild2, some samples may be missing labels in one of the three dimensions mentioned above. To address this, we employ the EMMA model\cite{li2022affective} to generate pseudo labels for these samples for subsequent training. 

\textbf{Evaluation Metrics}. The MTL evaluation metrics $\mathcal{P}_{mtl}$ consists of three parts: $\mathcal{P}_{au}, \mathcal{P}_{expr}, \mathcal{P}_{va} $. The specific formula is as follows:
\begin{align}
& \mathcal{P}_{au} = \frac{1}{12}\sum_{i=0}^{11} F_1^{au, i}, \\
& \mathcal{P}_{expr} = \frac{1}{8}\sum_{i=0}^{7} F_1^{expr, i}, \\
& \mathcal{P}_{va} =\frac{1}{2} (CCC_V + CCC_A), \\
& \mathcal{P}_{mtl} = \mathcal{P}_{au} + \mathcal{P}_{expr} + \mathcal{P}_{va} ,
\end{align}
where $F_1^{au, i}$ is the $F_1$ Score of 12 action units, $F_1^{expr, i}$ is the $F_1$ Score of 8 expressions, $CCC_V, CCC_A$ are the Concordance Correlation Coefficient (CCC) of valence and arousal, respectively.

\textbf{Implementation Details}. We directly choose to use the cropped-aligned version provided by the organizers. We resize all images from $ 112\times 112$ to $ 238\times 238$. Moreover, we also use data augmentation technique  during the training process. We take Adam\cite{kingma2014adam} as the optimizer. The initial learning rate is set to 0.001 and the cosine learning rate schedule is employed. The batch sizes is set to 64. The training epochs are 6, respectively, including 5 warm-up epochs. All the experiments are performed on an NVIDIA 4090 GPU.

\subsection{Results on the validation set}
 As shown in Tab. \ref{tab1}, we report the results of our model on the official validation set and compare it with the baseline(VGG-Face). Our best performance achieves overall score of \textbf{1.2542} on validation set. And it can be seen that the performance of our method far exceeds that of the baseline. Moreover, to avoid overfitting on the validation set, We performe 6-fold cross-validation experiment. Specifically, the training set is divided into 5 folds, and the validation set is considered as the 6th fold. The number of videos and frames in each fold is similar. The results are shown in Tab. \ref{tab2}.

\begin{table}[]
\centering
\footnotesize
\caption{The comparison between our model and baseline model on the official validation set.}
\begin{center}
\setlength{\tabcolsep}{4pt}
\begin{tabular}{lllll}
\hline
        & $\mathcal{P}_{au}$    & $\mathcal{P}_{expr}$  & $\mathcal{P}_{va}$    & $\mathcal{P}_{mtl}$   \\ \hline
Baseline   & - & - & - & 0.32 \\
Our model   & 0.4725 & 0.3484 & 0.4333 & 1.2542 \\ \hline
\end{tabular}
\label{tab1}
\end{center}
\end{table}

\begin{table}[]
\centering
\footnotesize
\caption{The results of the 6-fold cross-validation expreiments.}
\begin{center}
\setlength{\tabcolsep}{6pt}
\begin{tabular}{lllll}
\hline
        & $\mathcal{P}_{au}$    & $\mathcal{P}_{expr}$  & $\mathcal{P}_{va}$    & $\mathcal{P}_{mtl}$   \\ \hline
Fold1   & 0.5069 & 0.4301 & 0.3796 & 1.3166 \\
Fold2   & 0.4970 & 0.3528 & 0.5053 & 1.3551 \\
Fold3   & 0.4400 & 0.3617 & 0.3897 & 1.1913 \\
Fold4   & 0.5076 & 0.3824 & 0.5058 & 1.3959 \\
Fold5   & 0.4897 & 0.3432 & 0.4163 & 1.2492 \\
Fold6   & 0.4675 & 0.3358 & 0.4224 & 1.2257 \\ \hline
Average & 0.4848 & 0.3677 & 0.4365 & 1.2890 \\ \hline
\end{tabular}
\label{tab2}
\end{center}
\end{table}

\subsection{Results on the test set}
According to the rules, each team is allowed to have 5 different submissions. We will introduce our submission strategies. 
\begin{itemize}
\begin{sloppypar}
\item \textbf{1st submission}. We train models on the training set, and choose the models of epochs with the best overall performance (i.e. $\mathcal{P}_{mtl}$).
\item \textbf{2nd submission}. We train models on the training set, and choose the models of best epochs for different tasks. Specifically, we will obtain three models with different weights to handle each task, and then concatenate the predicted results for the 2nd submission.
\item \textbf{3rd submission}. We adopt the 6-fold strategy mentioned above for 3rd and 4th submissions, which requires conducting 6 sets of experiments. We select the model of epoch with the best overall performance(i.e. $\mathcal{P}_{mtl}$) from each set of experiments. This means that we will obtain six models with six settings, and all six models are used to ensemble to obtain the final result.
\item \textbf{4th submission}. For each task, we select the model from each set of experiments that performs the best. This means that for each subtask, there will be six models used to ensemble to obtain the final result of that task. Finally, the results of each task are concatenated to serve as the fourth submission. 
\item \textbf{5th submission}. The initial predicted results from the aforementioned four submission strategies are ensembled to obtain the final result, which serves as the fifth submission.
\end{sloppypar}
\end{itemize}

\section{Conclusion}
In this paper, we introduce our method for the Multi-Task Learning (MTL) Challenge of 4th AFFective Behavior Analysis in-the-wild (ABAW) competition.

\clearpage  

%
%
\bibliographystyle{splncs04}
\bibliography{main}
\end{document}